\documentclass[letterpaper]{article} 
\usepackage{aaai2026}  
\usepackage{times}  
\usepackage{helvet}  
\usepackage{courier}  
\usepackage[hyphens]{url}  
\usepackage{graphicx} 
\usepackage{natbib}  
\usepackage{caption} 
\usepackage{algorithm}
\usepackage{algorithmic}
\usepackage{xcolor}
\usepackage[table]{xcolor}
\usepackage{booktabs}
\usepackage[most]{tcolorbox}
\usepackage{subcaption}

\usepackage{newfloat}
\usepackage{listings}
\DeclareCaptionStyle{ruled}
{labelfont=normalfont,labelsep=colon,strut=off} 
\floatstyle{ruled}
\newfloat{listing}{tb}{lst}{}
\floatname{listing}{Listing}
\title{Revisiting MLLM Based Image Quality Assessment: Errors and Remedy}
\author{
    Zhenchen Tang\textsuperscript{\rm 1,2},
    Songlin Yang\textsuperscript{\rm 3},
    Bo Peng\textsuperscript{\rm 1}\equalcontrib,
    Zichuan Wang\textsuperscript{\rm 1,2},
    Jing Dong\textsuperscript{\rm 1}\equalcontrib
}
\affiliations{
    \textsuperscript{\rm 1}New Laboratory of Pattern Recognition, Institute of Automation, Chinese Academy of Sciences\\
    \textsuperscript{\rm 2}School of Artificial Intelligence, University of Chinese Academy of Sciences\\
    \textsuperscript{\rm 3}MMLab@HKUST, The Hong Kong University of Science and Technology

    \{tangzhenchen2024, wangzichuan2024\}@ia.ac.cn, syangds@connect.ust.hk,
    \{bo.peng, jdong\}@nlpr.ia.ac.cn
}

\usepackage{bibentry}

\begin{document}

\maketitle

\begin{abstract}
The rapid progress of multi-modal large language models (MLLMs) has boosted the task of image quality assessment (IQA).
However, a key challenge arises from the inherent mismatch between the discrete token outputs of MLLMs and the continuous nature of quality scores required by IQA tasks. This discrepancy significantly hinders the performance of MLLM-based IQA methods. Previous approaches that convert discrete token predictions into continuous scores often suffer from conversion errors. Moreover, the semantic confusion introduced by level tokens (e.g., “good”) further constrains the performance of MLLMs on IQA tasks and degrades their original capabilities to related tasks.
To tackle these problems, we provide a theoretical analysis of the errors inherent in previous approaches and, motivated by this analysis, propose a simple yet effective framework, Q-Scorer. This framework incorporates a lightweight regression module and IQA-specific score tokens into the MLLM pipeline. Extensive experiments demonstrate that Q-Scorer achieves state-of-the-art performance across multiple IQA benchmarks, generalizes well to mixed datasets, and further improves combined with other methods.

\end{abstract}

\begin{links}
    \link{Code}{https://github.com/2kxx/Q-Scorer}
\end{links}

\section{Introduction}

Image Quality Assessment (IQA) is a fundamental task in computer vision, aiming to quantify the perceptual quality of images in a way that aligns closely with human visual perception. With the recent advances in reinforcement learning-based post-training paradigms, IQA has become increasingly important as a reward signal for downstream tasks, such as image generation~\cite{liang2024rich, yu2024sf,li2025beyond,li2025instant,hanadaptive} and enhancement~\cite{zheng2021learning, zhou2022quality,wang2025handeval}.

\begin{figure}[t]
\centering
\includegraphics[width=0.45\textwidth]{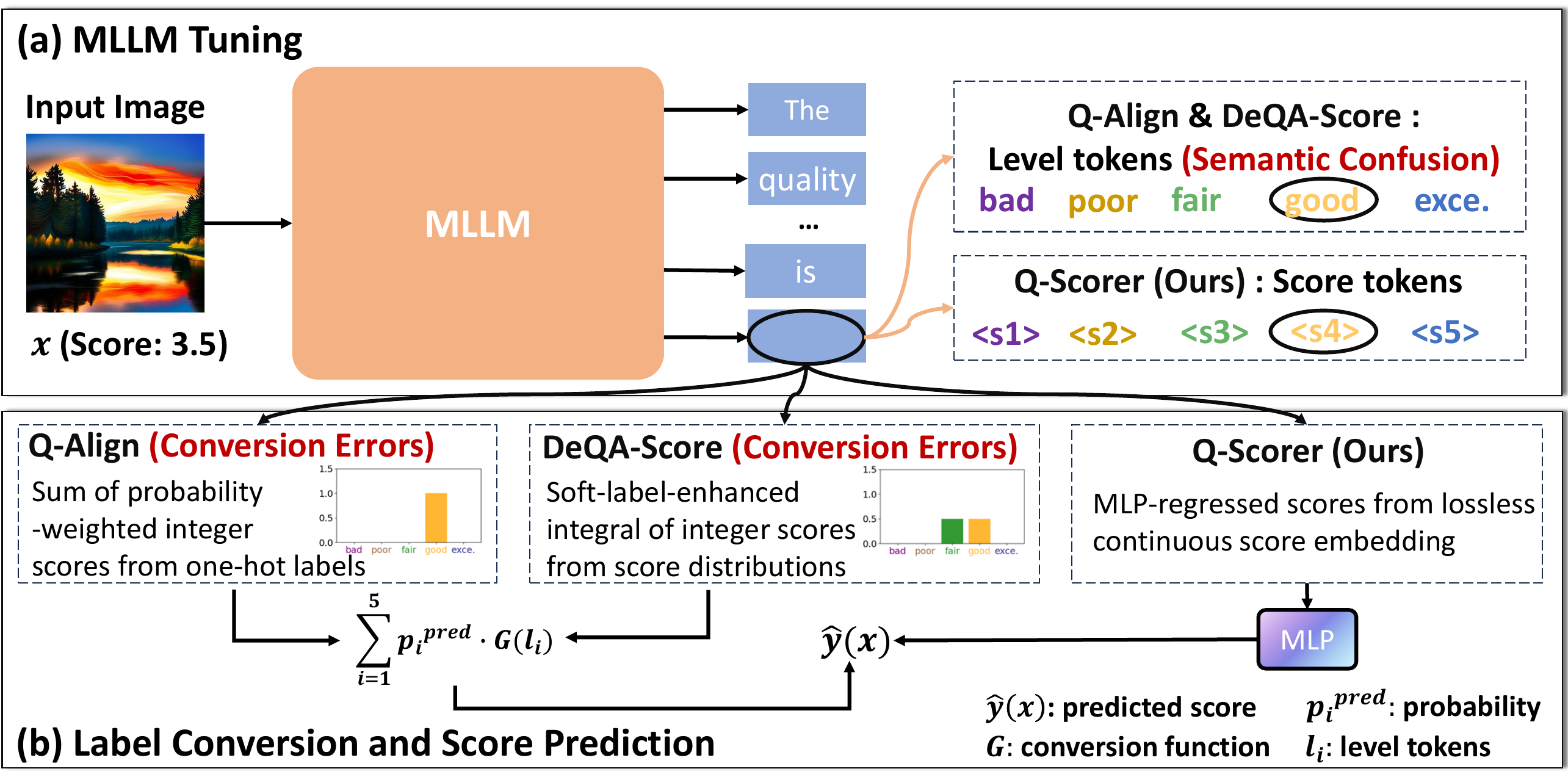}

\caption{Overview of MLLM-based IQA methods and error analysis. The figure shows how MLLMs are adapted for label conversion and score prediction, and highlights steps causing conversion errors and semantic confusion.}
\label{fig1}
\vspace{-0.4cm}
\end{figure}

Recently, leveraging MLLMs~\cite{liu2023visual, ye2024mplug, bai2025qwen2, chen2024expanding} to improve IQA performance has shown promise, as MLLMs can understand visual images using linguistic intelligence, which greatly facilitates aligning model predictions with human evaluations across diverse image types. However, existing MLLM-based IQA methods~\cite{wu2023qb, wu2023qa, wu2024q} typically formulate the task as learning the correlation between images and discrete textual tokens (e.g., “poor,” “good”), which fits the token output nature of MLLMs but conflicts with the intrinsic nature of IQA — predicting continuous quality scores. As shown in Fig.\ref{fig1}, although several studies~\cite{wu2023qa, you2025teaching, li2025tokenenoughrealisticimage} has explored techniques to bridge this mismatch, two key problems remain:



(1) Conversion Errors: Existing token-based approaches suffer from inevitable theoretical errors during the label conversion process. Since real-valued Mean Opinion Scores (MOSs) are quantized into discrete tokens, even perfect classification cannot exactly recover the ground-truth MOS. We theoretically analyze this issue and prove that such errors are intrinsic to token-based IQA formulations in Sec.~\ref{conversion errors}. 


(2) Semantic Confusion: Most quality-related textual descriptions (i.e., level tokens) used in current methods are drawn from pre-trained LLM vocabularies, whose semantics are not optimized for IQA. This introduces two problems: pre-tuning semantic confusion, where these tokens carry prior meanings unrelated to visual quality; post-tuning semantic confusion, where training on IQA-specific tasks may overwrite the semantic integrity of these tokens.

To address these problems, we propose \textbf{Q-Scorer}, a simple yet effective framework that incorporates a lightweight regression module and IQA-specific score tokens into the MLLM pipeline. For conversion errors, we adopt an MLP-based module to regress the continuous IQA scores. Although prior work used a linear layer on the final token~\cite{he2024videoscore}, its entanglement with IQA-irrelevant information limits performance compared to token-based methods (even the baseline Q-Align), which causes insufficient embedding expressiveness. To address this issue and mitigate semantic confusion, we introduce a set of IQA-specific score tokens. These tokens guide the MLP module to improve IQA score prediction without compromising the original capabilities of MLLMs.


Equipped with these designs, our method achieves state-of-the-art performance using only LoRA-based fine-tuning on limited number of model parameters. Q-Scorer consistently outperforms baseline approaches in score regression, and despite being trained solely on the KonIQ dataset, it also exhibits strong generalization to other datasets. Furthermore, it generalizes well to mixed-dataset settings and can be seamlessly integrated with other methods to achieve even better performance.

Our contributions can be summarized as follows:
\begin{itemize}
\item We conduct error analysis and find that existing MLLM-based IQA methods suffer from conversion errors and semantic confusion when using discrete level tokens.

\item To address these challenges, we propose Q-Scorer, a simple yet effective framework that leverages IQA-specific score tokens and an MLP regressor.

\item Extensive experiments demonstrate that our framework achieves state-of-the-art performance across multiple benchmarks, generalizes well to mixed datasets, and further improves combined with other methods.
\end{itemize}

\section{Related Works}
\subsection{Traditional IQA Methods}
Prior to the emergence of MLLMs, traditional IQA methods were generally categorized into full-reference (FR) and no-reference (NR) approaches. Classical FR methods rely on handcrafted similarity metrics that compare a distorted image with its high-quality reference, such as PSNR~\cite{hore2010image} and SSIM~\cite{wang2004image}. In contrast, NR methods estimate perceptual quality directly from a single image using natural scene statistics, as exemplified by BRISQUE~\cite{mittal2012no} and NIQE~\cite{mittal2012making}. With the advancement of deep learning, data-driven IQA approaches have achieved significant improvements by directly regressing human perceptual scores from raw image inputs~\cite{bosse2017deep, talebi2018nima, su2020blindly}. More recent studies further boost performance and generalization by incorporating multi-scale features~\cite{ke2021musiq}, co-training on multiple datasets~\cite{zhang2021uncertainty}, multitask learning~\cite{zhang2023blind}, or leveraging pretrained vision-language models such as CLIP~\cite{wang2023exploring, tang2024clip}. 

\subsection{MLLM-Based IQA methods}
MLLM-based IQA methods leverage the foundational knowledge embedded in MLLMs to improve IQA performance and enhance generalization. Q-Bench~\cite{wu2023qb} demonstrates that MLLMs possess low-level visual perception and understanding, enabling them to predict quantifiable quality scores. Q-Instruct~\cite{wu2024q} improves question-answering accuracy including IQA tasks, by introducing instruction-response style description dataset. DepictQA~\cite{you2024depicting} and its extension~\cite{you2024descriptive} generates detailed quality assessments by training on large-scale explanation dataset. Moreover, Q-Ground~\cite{chen2024qg} and SEAGULL~\cite{chen2024seagull} extend IQA to more fine-grained tasks, including visual quality grounding and region-aware scoring.

\subsection{IQA Scorers}
Although numerous MLLM-based IQA methods have emerged recently, their scoring strategies are limited and can only be categorized into \textbf{token-based} and \textbf{regression-based approaches}, depending on whether they utilize an MLP~\cite{li2025tokenenoughrealisticimage}. We refer to the methods proposing these scoring strategies as IQA Scorers. First are token-based methods. Initially, Q-Bench~\cite{wu2023qb} introduces a binary softmax strategy to predict quality scores by two discrete levels. Compare2Score~\cite{zhu2024adaptive} adopts a pairwise comparison strategy to infer image quality scores. Building on Q-Bench, the widely adopted method Q-Align~\cite{wu2023qa} improves scoring process by discretizing quality into five levels and training MLLMs using one-hot labels. DeQA-Score~\cite{you2025teaching} further enhances Q-Align by introducing soft labels based on score distributions. RealQA~\cite{li2025tokenenoughrealisticimage} takes a different approach by directly predicting two extra significant digits of the score. 

In contrast, regression-based methods are rarely explored in MLLM-based IQA. While MLPs are commonly used in traditional IQA models, their application within MLLMs remains limited due to the dominance of the next-token prediction paradigm, which makes it challenging to integrate continuous regression objectives effectively. Among the few existing attempts, VideoScore~\cite{he2024videoscore} performs direct numerical regression via a linear output layer. However, its insufficient embedding expressiveness limits its performance compared to token-based methods.

\begin{figure}[t]
\centering
\includegraphics[width=0.45\textwidth]{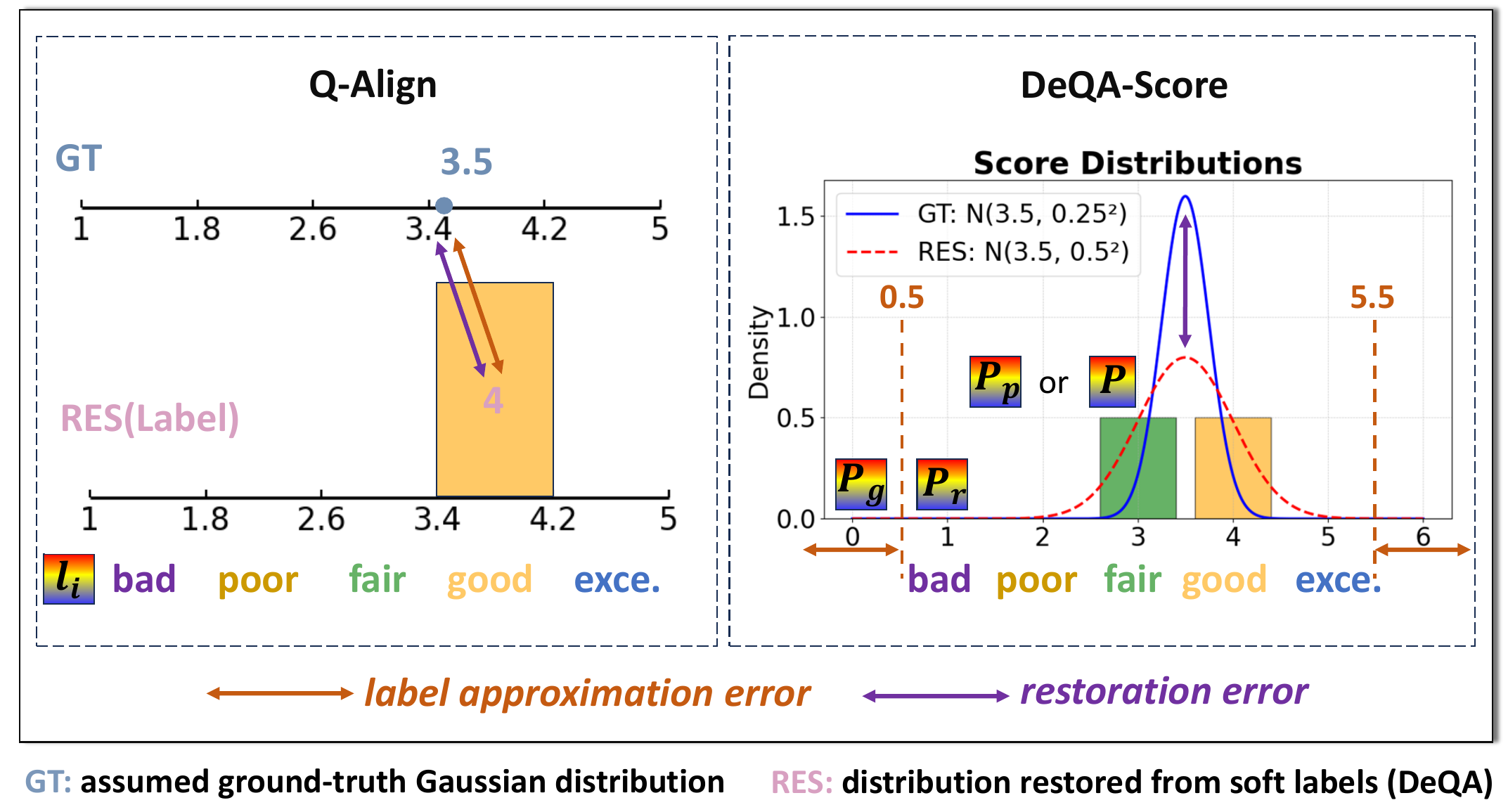}
\caption{Visual illustration and detailed explanation of conversion errors. The figure provides examples detailing two main sources of conversion errors: label approximation (from discretizing MOS) and restoration error (from imperfect score restoration). See Sec.~\ref{semantic confusion} and Fig.\ref{fig3} for details.}
\label{fig2}
\vspace{-0.4cm}
\end{figure}

\section{Error Analysis: Conversion Errors and Semantic Confusion}
\label{error analysis}
Previous methods for formulating MLLM-based IQA typically adopt discrete level tokens (i.e., ``bad", ``poor", ``fair", ``good", ``excellent") to describe image quality. However, this level-token-based formulation introduces two potential sources of errors: conversion errors (Sec.~\ref{conversion errors}) and semantic confusion (Sec.~\ref{semantic confusion}). For conversion errors, we analyze the limitations of discrete token-based score fitting and explain why modeling IQA as a regression task can mitigate this issue. For semantic confusion, we present insights from both pre-tuning and post-tuning perspectives, highlighting the problems arising from directly using level tokens.

\begin{algorithm}[t]
\caption{\textbf{Q-Align}}
\label{alg:qalign}
\begin{algorithmic}[1]
\STATE \textbf{Input:} Image $x$
\STATE \textbf{Target:} Normalized ground-truth MOS $S \in [1, 5]$
\STATE \textbf{Output:} Predicted quality score $\hat{y}(x)$

\STATE \textbf{Training Phase:}
\STATE Divide the range $[1,5]$ into 5 intervals $I_1, I_2, \dots, I_5$
\FOR{$i = 1$ to $5$}
    \IF{$1 + \frac{4}{5}(i - 1) < S \leq 1 + \frac{4}{5}i$}
        \STATE Assign label $L(S) \gets l_i$
    \ENDIF
\ENDFOR

\STATE \textbf{Level tokens:} $\{l_i\} = \{\textit{bad}, \textit{poor}, \textit{fair}, \textit{good}, \textit{excellent}\}$

\STATE \textbf{Inference Phase:}
\STATE Predict probability $p_i^{pred}$ for each level token $l_i$
\STATE $G(l_i)$ convert $l_i$ to discrete score $i$
\STATE Predict final score: $\hat{y}(x) = \sum_{i=1}^{5} p_i^{pred} \cdot G(l_i)$

\RETURN $\hat{y}(x)$
\end{algorithmic}
\end{algorithm}

\subsection{Conversion Errors}
\label{conversion errors}
To align with the discrete token outputs of MLLMs, ground-truth IQA scores are typically quantized into class-level labels. However, as illustrated in Fig.\ref{fig2}, this design introduces two types of conversion errors: \textbf{label approximation error} and \textbf{restoration error}. The label approximation error, which arises from converting a continuous ground-truth MOS $S$ into a discrete label that MLLMs can process. The restoration error, referring to the discrepancy between the predicted score $\hat{y}(x)$ (where $x$ is an input image) and the ground-truth MOS $S$ due to imperfect restoration from the learned label representation. In this section, we first formulate the errors of representative methods and then analyze why regression-based modeling can effectively mitigate conversion errors. The proof details are provided in Appendix.


\noindent \textbf{(a) Error Formulation: Q-Align and DeQA-Score}

\noindent \textbf{Q-Align} serves as the foundational paradigm for most current MLLM-based IQA methods. As shown in Algorithm~\ref{alg:qalign}, it learns from one-hot score labels and predicts the final score using a weighted sum of integer scores from level token probabilities. However, this process involves only discrete textual supervision, leading to the loss of continuous score information during training.

Assuming that the model makes a perfectly accurate prediction (i.e., assigning probability 1 to the correct discrete label), the expected predicted score $\hat{y}(x)$ will be $G(l_j)$, where $j$ denotes the index of the predicted level token. In Fig.\ref{fig2}, the model predicts level token $l_4$, and the resulting score is $\hat{y}(x) = G(l_4) = 4$. In this case, the label approximation error can be regarded as equivalent to the restoration error. The one-sample error $\epsilon(x)$ can thus be calculated as:
\begin{equation}
\epsilon(x) = |G(l_j) - S| \ge 0.
\end{equation}

Under this assumption, we treat the error $\epsilon(x)$ as uniformly distributed within the interval. Therefore, the expected theoretical error $\mathrm{E}[\epsilon(x)^2] = \frac{18}{125} > 0$. 

\begin{algorithm}[t]
\caption{\textbf{DeQA-Score}}
\label{algo:deqa}
\begin{algorithmic}[1]
\STATE \textbf{Input:} Image $x$
\STATE \textbf{Target:} Assume $S$ follows Gaussian distribution (GT): $s \sim \mathcal{N}(\mu, \sigma^2)$.
MOS as $\mu$ and annotated variance as $\sigma^2$
\STATE \textbf{Output:} Predicted quality score $\hat{y}(x)$
\STATE \textbf{Training Phase:}
\STATE Divide the range $[0.5,5.5]$ into 5 intervals $I_1^{\prime}, I_2^{\prime}, \dots, I_5^{\prime}$
\STATE Define soft label $P_r = \{p_1^{raw}, p_2^{raw}, p_3^{raw}, p_4^{raw}, p_5^{raw}\}$ 
\STATE Define interval midpoint $c_i \in \{1,2,3,4,5\}$ 
\FOR{$i = 1$ to $5$}
    \STATE $p_i^{raw} = \int_{c_i - \frac{1}{2}}^{c_i + \frac{1}{2}} f(s)\, ds$
\ENDFOR
\STATE \textit{Enhancing soft label accuracy:}
\STATE Linearly transform $P_r$ to enhanced soft lable $P$:
\STATE $\quad p_i = \alpha \cdot p_i^{raw} + \beta$, with $\sum_i p_i = 1$, $\sum_i p_i c_i = \mu$
\STATE Use model to predict level token distribution $P_p$
\STATE $P_p = \{p_1^{pred}, p_2^{pred}, p_3^{pred}, p_4^{pred}, p_5^{pred}\}$
\STATE Minimize KL divergence: $\mathcal{L}_{KL}(P \| P_p)$
\STATE \textbf{Inference Phase:}
\STATE Predict final score: $\hat{y}(x) = \sum_{i=1}^{5} p_i^{pred}(x) \cdot c_i$
\RETURN $\hat{y}(x)$
\end{algorithmic}
\end{algorithm}

\begin{figure*}[t]
\centering
\includegraphics[width=0.45\textwidth]{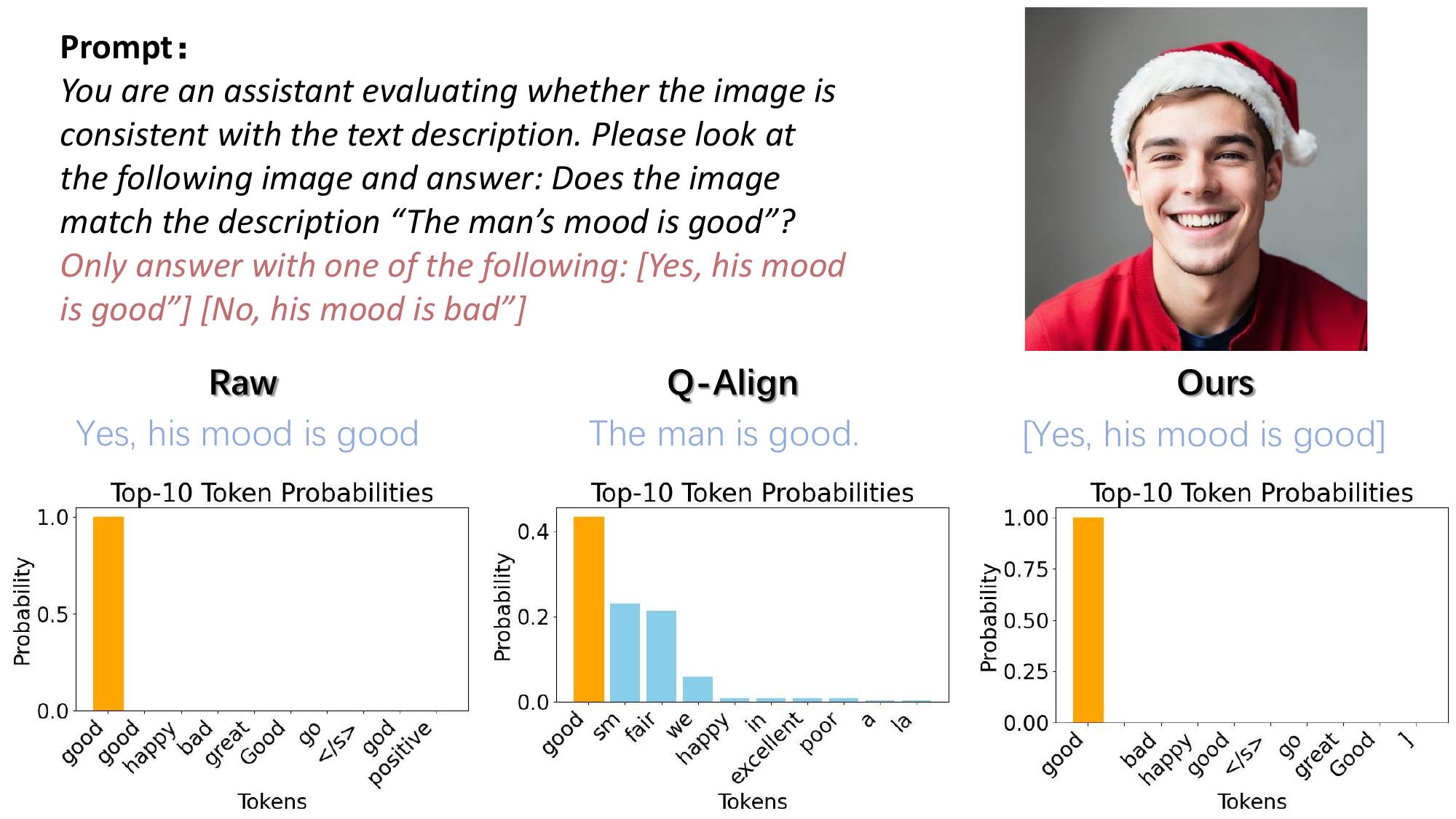}
\hspace{0.01\textwidth}
\includegraphics[width=0.45\textwidth]{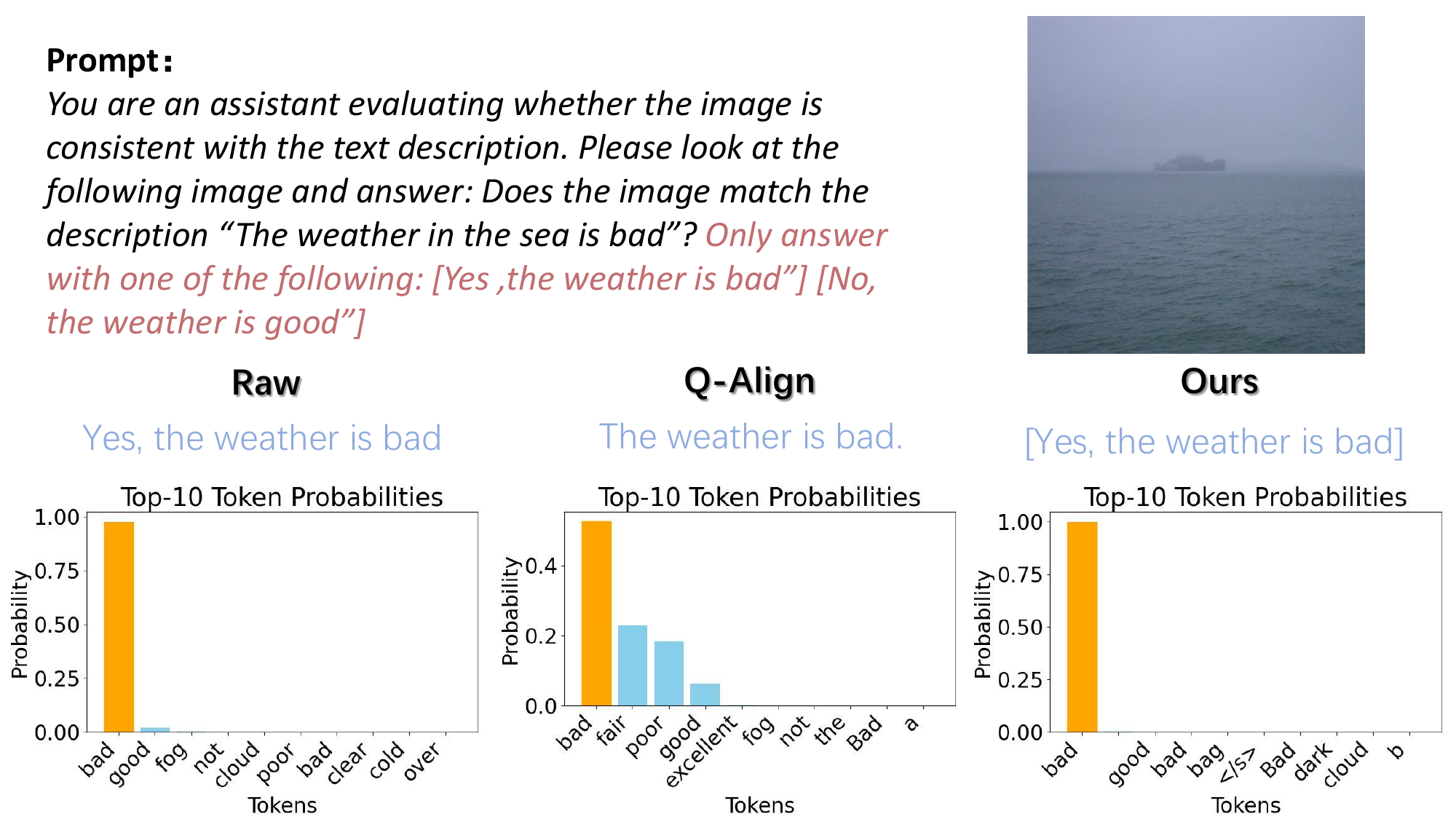}
\caption{Examples of post-tuning semantic confusion, showing how different token strategies affect T2I alignment assessment.}
\label{fig3}
\vspace{-0.4cm}
\end{figure*}

\noindent \textbf{DeQA-Score}, as shown in Algorithm~\ref{algo:deqa}, learns from soft-label-enhanced integral of integer scores from the score distribution and predicts the final score via the same probability-weighted summation as Q-Align, which is proposed to mitigate the label approximation error in Q-Align.

However, this formulation introduces certain theoretical errors. Assuming the model makes a perfectly accurate prediction, two sources of error remain: (1) The Gaussian prior itself is an approximation of the true, human rating distribution. This modeling discrepancy leads to an inherent systematic error $\epsilon_1(x)$. (2) The interval-based truncation leads to $\sum_i p_i^{raw} < 1$, which means that raw soft label $P_{r}$ does not form a strictly valid discrete probability distribution. 

As a result, the label and the predicted score $\hat{y}(x)$ inevitably deviate from the ground-truth MOS $S$, leading to a label approximation error $\epsilon_2(x)$:
\begin{equation}
\epsilon_2(x) = \left| \hat{y}(x) - S \right| 
= \left| \sum_{i=1}^{5} p_i^{raw} c_i - \int_{-\infty}^{\infty} sf(s)\, ds \right| > 0,
\end{equation}

Although DeQA-Score applies a linear transformation to refine $P_{r}$ into enhanced soft label $P$, theoretically reducing the label approximation error to zero. As shown in Fig.\ref{fig2}, the process of using the discrete distribution $P$ to recover the original continuous Gaussian distribution $P_g$ (GT) inevitably causes a restoration error:
\begin{equation}
\mu^{res} = \sum\nolimits p_i^{pred} c_i, \quad 
(\sigma^{res})^2 = \sum\nolimits p_i^{pred} (c_i - \mu^{res})^2
\end{equation}
This error is severe when the original variance is small, resulting in a significant increase in error $\epsilon_1$ ~\cite{you2025teaching}.

Combining these, the error is always greater than zero:
\begin{equation}
\epsilon(x) = \epsilon_1(x) + \epsilon_2(x) > 0,
\end{equation}

Assuming $\epsilon_1$ and $\epsilon_2$ are independent, the total theoretical error is always greater than zero:
\begin{equation}
\mathrm{E}[\epsilon(x)^2] = \mathrm{E}[\epsilon_1^2] + \mathrm{E}[\epsilon_2^2] > 0.
\end{equation}

\noindent \textbf{(b) Error Mitigation: Regression-Based Modeling}

Given the errors discussed above, we explore regression-based modeling as a mitigation. In regression-based models, the predicted score $\hat{y}(x)$ is obtained via a regression function $\mathcal{F}(x; \theta)$ over the input feature $x$:
\begin{equation}
\hat{y}(x) = \mathcal{F}(x; \theta).
\end{equation}

Since regression-based models use the ground-truth MOS $S$ directly as supervision without discretization, their theoretical label approximation error is inherently zero. Therefore, only the restoration error needs to be considered. Let $g(x)$ denote the function mapping input $x$ to the corresponding MOS $S$. Then, for any $\varepsilon > 0$, there exists a feedforward neural network with a single hidden layer of the form:
\begin{equation}
\mathcal{F}(x) = \sum_{i=1}^n \alpha_i \cdot \sigma(w_i^\top x + b_i),
\end{equation}
where $\sigma(\cdot)$ is an activation function (e.g., Sigmoid or ReLU), and $\{\alpha_i, w_i, b_i\}_{i=1}^n$ are the learnable parameters.

By Universal Approximation Theorem (UAT)~\cite{hornik1989multilayer}, such a network satisfies:
\begin{equation}
\epsilon(x) = \sup_{x} \left| \mathcal{F}(x) - g(x) \right| = \sup_{x} \left| \hat{y}(x) - S \right| < \varepsilon,
\end{equation}
The uniform approximation also implies the expected error:
\begin{equation}
\mathrm{E}[\epsilon(x)^2] = \inf_{\theta} \mathrm{E}_x \left[\left( \hat{y}(x) - S \right)^2 \right] < \varepsilon^2,
\label{eq:uat_mse_bound}
\end{equation}

As $\varepsilon \to 0$, the bound vanishes. We obtain $\mathrm{E}[\epsilon(x)^2] = 0$, indicating that with sufficiently many model parameters, the theoretical error of regression-based methods can be reduced to zero, whereas the error of previous token-based methods remains strictly positive. As for RealQA~\cite{li2025tokenenoughrealisticimage}, it directly outputs scores with a maximum valid digit of two, which inherently introduces restoration error.\par

This result guarantees that, with sufficiently rich embeddings $x$ and enough hidden units $\theta$, an MLP can theoretically approximate the ground-truth MOS function $g(x)$ arbitrarily well. This continuous mapping naturally offers finer granularity and better capacity to represent real-valued scores.

\begin{table}[t]
\centering
\setlength{\tabcolsep}{4pt}
\footnotesize
\begin{tabular}{llc}
\toprule
\textbf{Category} & \textbf{Level Texts} & \textbf{KonIQ} \\
\midrule
Common & excellent / good / fair / poor / bad & 0.932 / 0.938 \\
Reverse & bad / poor / fair / good / excellent & 0.922 / 0.931 \\
Number & five / four / three / two / one & 0.924 / 0.934 \\
Random & apple / fog / tree / bag / car & 0.925 / 0.936 \\
\rowcolor{gray!40} 
\textbf{Ours} & \textless s5\textgreater / \textless s4\textgreater / \textless s3\textgreater / \textless s2\textgreater / \textless s1\textgreater & \textbf{0.944 / 0.942} \\

\bottomrule
\end{tabular}
\caption{Comparison of different token settings on KonIQ. “Ours” replaces level tokens with score tokens, while other settings remain consistent with Q-Align. (PLCC / SRCC)}
\label{tab1}
\vspace{-0.4cm}
\end{table}

\subsection{Semantic Confusion}
\label{semantic confusion}

Token-based IQA methods often directly use existing vocabulary tokens to represent quality levels. However, this introduces a phenomenon we refer to as semantic confusion, which occurs \textbf{Pre-tuning} and \textbf{Post-tuning}. It fundamentally limits the effectiveness and transferability of such methods.

\subsubsection{Pre-Tuning Confusion.} This arises because MLLMs already associate certain tokens with inherent semantic meanings. For example, words such as “good” or “bad”, beyond representing quality, are strongly associated with general sentiment and weather conditions. When these tokens are only reused to represent discrete quality levels in IQA tasks, the pre-trained semantics can interfere with accurate quality judgment. As shown in Tab.\ref{tab1}, using level tokens from Q-Align (Common) underperforms our method due to prior meanings unrelated to visual quality. Moreover, reversing semantic alignment (e.g., mapping “good” to “poor”) significantly degrades performance. Both highlight the impact of pre-trained semantics. Similarly, using numeric or random tokens can also introduce noise into the learning process.

\subsubsection{Post-Tuning Confusion.} This refers to the fact that fine-tuning quality-related tokens for IQA can overwrite their semantic integrity and eventually disrupt their use in other tasks. In Fig.\ref{fig3}, we show that after Q-Align training, the model’s ability to assess even related tasks like text-to-image (T2I) alignment using these tokens is noticeably impaired. The model tends to overgenerate the IQA-specific sentence pattern (e.g., “The xxx is...”), and the output probability of all level tokens increases sharply, even when unrelated to the given prompt. To mitigate both forms of semantic confusion, we introduce IQA-specific tokens that do not carry pre-existing semantic interference. As evidenced in Tab.\ref{tab1} and Fig.\ref{fig3}, our proposed tokens not only achieve better performance than level tokens in the IQA task, but also preserve the model's capabilities in other tasks.

\section{Error Remedy: Q-Scorer}
To alleviate the errors discussed in Sec.~\ref{error analysis}, we propose our simple yet effective solution, Q-Scorer. 


\subsection{Model Architecture}
\label{model architecture}
\subsubsection{Visual Encoder and Abstractor.}
\begin{figure}[t]
\centering
\includegraphics[width=0.45\textwidth]{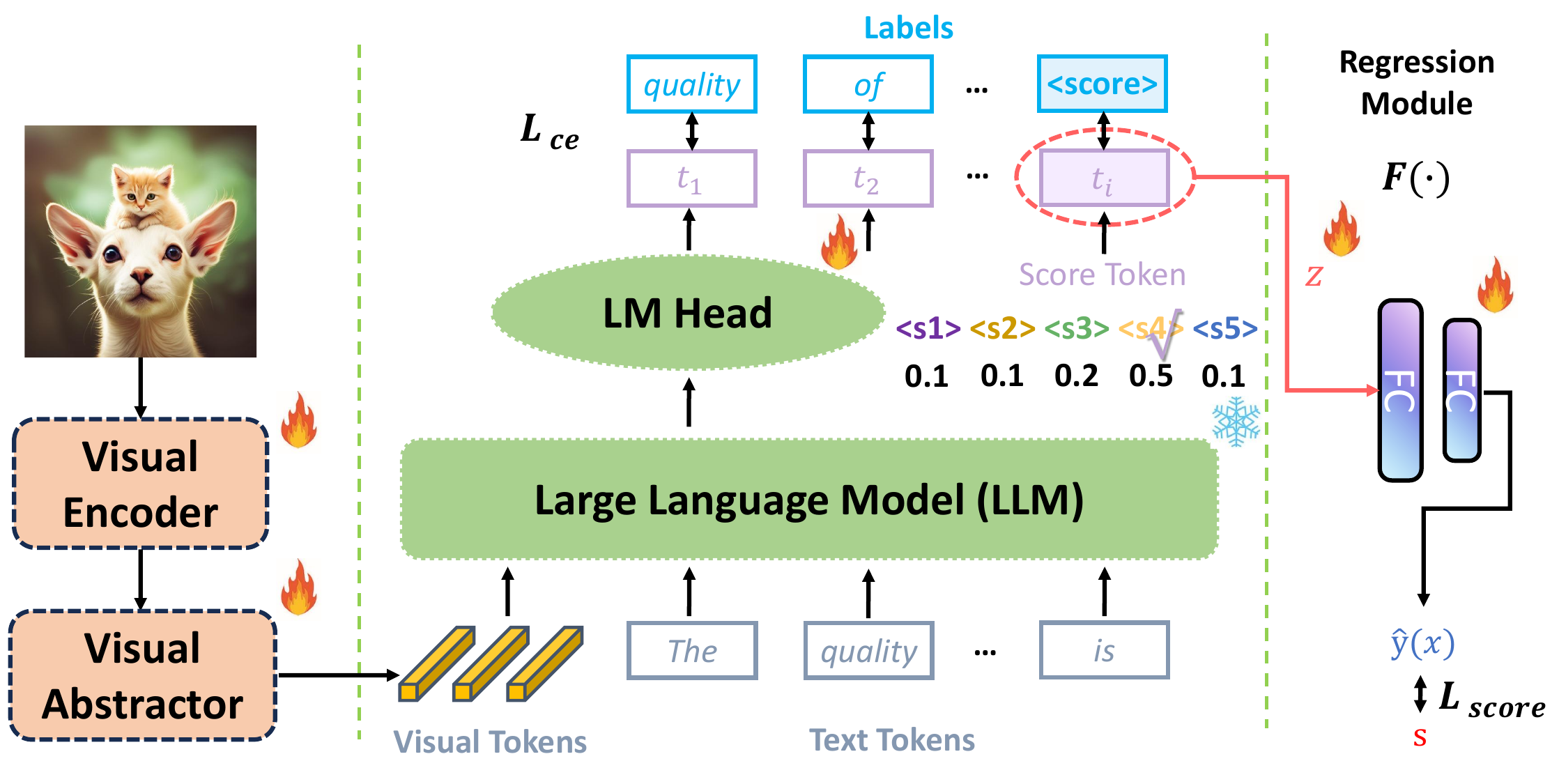}
\caption{Overview of Q-Scorer. It uses $\mathcal{L}_{\text{ce}}$ learn to output an interval‑specific score token. The token's embedding is then passed to an MLP to regress the continuous quality score, optimized with $\mathcal{L}_{\text{score}}$ to preserve the lossless MOS.}
\label{fig4}
\vspace{-0.4cm}
\end{figure}

As illustrated in Fig.\ref{fig4}, we adopt mPLUG-Owl2~\cite{ye2024mplug} as the base architecture to construct our model. We inherit both the visual encoder and the visual abstractor modules from mPLUG-Owl2 to process visual information. Specifically, the input image is first encoded into 1024 visual tokens via the visual encoder. Then, a visual abstractor is applied to compress these tokens into a more compact set of 64 tokens. Finally, reduced visual tokens are fused with prompt-generated text tokens and fed jointly into the LLM for response generation.\par

\subsubsection{IQA-Specific Score Token.} 
Although VideoScore~\cite{he2024videoscore} employs a regression module, experimental results show that due to the limited representation capacity of the embedding $x$, its performance is inferior even compared to token-based methods that suffer from errors. To address the aforementioned issues of semantic confusion and insufficient embedding expressiveness in regression-based IQA scoring, we introduce a set of IQA-specific score tokens directly into the language model’s vocabulary.

While adding a special token has been effective in many MLLM-based downstream tasks, using a single score token $\{score\}$ often leads to highly similar embeddings across different inputs, because the model is trained to consistently produce the same token regardless of input variations. This makes it difficult for a shallow MLP to regress accurate scores from the similar embeddings. To mitigate this, we define a group of discrete tokens $\{score1, score2, score3, score4, score5\}$, each corresponding to a specific score interval, which mirrors the approach in Q-Align. During training, the model learns to select the appropriate score token \texttt{<scorex>} based on the quality level, and a lightweight MLP regressor predicts a fine-grained score within that interval by this score token. This approach reduces semantic interference from existing vocabulary and enhances the model’s ability to distinguish between fine-level quality differences, enabling more accurate score prediction.\par

\subsubsection{Regression Module.}
Motivated by the theoretical conversion errors discussed earlier, we design a lightweight regression module that directly predicts image quality scores from score token embeddings. Specifically, we extract the embedding of the corresponding IQA-specific score token, which encodes both visual and quality-related semantics, and feed it into an MLP with a $4096\rightarrow2048\rightarrow1024\rightarrow1$ architecture to produce a scalar quality prediction. This direct regression strategy eliminates the need for token-to-score conversions in token-based methods, which are often lossy and irreversible, effectively narrowing the gap between predicted scores and human perception.

\subsection{Training Loss}
\label{training loss}
We use the standard cross-entropy loss $L_{ce}$ to train the model to predict the correct score token. Following typical LLM training~\cite{touvron2023llama}, we compute the next-token prediction loss as:

\begin{equation}
L_{ce} = - \sum_{i=1}^{T} \log P(t_i \mid t_1, t_2, \ldots, t_{i-1})
\end{equation}
where $T$ is the sequence length, $t_i$ is the $i$-th token, and $P(t_i \mid t_1,\ldots,t_{i-1})$ denotes the probability of predicting token $t_i$ given the preceding tokens.

In addition, we extract the embedding $z$ of the predicted score token \texttt{<scorex>} and use an MLP $F(\cdot)$ to predict the final score. We apply a MSE loss between the predicted score and ground-truth MOS:

\begin{equation}
L_{score} = (F(z) - S)^2
\end{equation}
This dual-loss design enables continuous MOS supervision to be seamlessly embedded into discrete score tokens, achieving lossless label within the MLLM framework.

\subsection{Conversation Formats}
\label{conversation formats}
We define a concise conversation format tailored for the IQA task. Let the image token be \texttt{<img>} and the predicted score token be \texttt{<scorex>}. The dialogue format is as follows:

\begin{flushleft}
\#User: \texttt{<img>} How would you rate the quality of this image? \\
\#Assistant: The quality of this image is \texttt{<scorex>}.
\end{flushleft}

\begin{table*}[htbp]
\centering
\footnotesize
\begin{tabular}{l|l|c|ccccc}
\toprule
\textbf{Category} & \textbf{Methods} & \textbf{KonIQ} & \textbf{SPAQ$^{cr}$} & \textbf{KADID$^{cr}$}  & \textbf{LIVE-Wild$^{cr}$} & \textbf{AGIQA-3K$^{cr}$} & \textbf{CSIQ$^{cr}$} \\
\midrule
Traditional
& NIQE & 0.533 / 0.530 & 0.679 / 0.664 & 0.468 / 0.405 & 0.493 / 0.449 & 0.560 / 0.533 & 0.718 / 0.628 \\
& BRISQUE & 0.225 / 0.226 & 0.490 / 0.406 & 0.490 / 0.406 & 0.361 / 0.313 & 0.541 / 0.497 & 0.740 / 0.556 \\
\midrule
Learning-based
& NIMA & 0.896 / 0.859 & 0.838 / 0.856 & 0.532 / 0.535 & 0.814 / 0.771 & 0.715 / 0.654 & 0.695 / 0.649 \\
& HyperIQA & 0.917 / 0.906 & 0.791 / 0.788 & 0.506 / 0.468 & 0.772 / 0.749 & 0.702 / 0.640 & 0.752 / 0.717 \\
& DBCNN & 0.884 / 0.875 & 0.812 / 0.806 & 0.497 / 0.484 & 0.773 / 0.755 & 0.730 / 0.641 & 0.586 / 0.572 \\
& MUSIQ & 0.924 / 0.929 & 0.868 / 0.863 & 0.575 / 0.556 & 0.789 / 0.830 & 0.722 / 0.630 & 0.771 / 0.710 \\
& CLIP-IQA+ & 0.909 / 0.895 & 0.866 / 0.864 & 0.653 / 0.654 & 0.832 / 0.805 & 0.736 / 0.685 & 0.772 / 0.719 \\
& ManIQA & 0.849 / 0.834 & 0.768 / 0.758 & 0.499 / 0.465 & 0.849 / 0.832 & 0.723 / 0.636 & 0.623 / 0.627 \\
\midrule
MLLM-based
& Compare2Score & 0.923 / 0.910 & 0.867 / 0.860 & 0.500 / 0.453 & 0.786 / 0.772 & 0.777 / 0.671 & 0.735 / 0.705 \\
& Q-Align & 0.941 / 0.940 & 0.886 / 0.887 & 0.674 / 0.684 & 0.853 / 0.860 & 0.772 / 0.735 & 0.785 / 0.737 \\
& DeQA-Score & 0.953 / 0.941 & 0.895 / 0.896 & \textbf{0.694 / 0.687} & 0.892 / 0.879 & 0.809 / 0.729 & 0.787 / 0.744 \\
& Q-Align (LoRA) & 0.932 / 0.938 & 0.874 / 0.886 & 0.624 / 0.632 & 0.858 / 0.859 & 0.806 / 0.735 & 0.772 / 0.730 \\
\rowcolor{gray!30} 
& \textbf{Ours (5)} & 0.959 / 0.948 & 0.898 / 0.898 & 0.676 / 0.671 & 0.889 / 0.870 & \textbf{0.821 / 0.736} & \textbf{0.796 / 0.746} \\
\rowcolor{gray!50} 
& \textbf{Ours (1)} & \textbf{0.960 / 0.950} & \textbf{0.900 / 0.899} & 0.660 / 0.645 & \textbf{0.903 / 0.888} & 0.811 / 0.722 & 0.795 / 0.733 \\
\bottomrule
\end{tabular}
\caption{Performance comparison across multiple IQA datasets. “cr” denotes cross-dataset evaluation. “Ours(1)” refers to predicting the same score token regardless of input. “Ours(5)” denotes our base method. (PLCC / SRCC)}
\label{tab2}
\vspace{-0.3cm}
\end{table*}

\begin{table}[htbp]
\centering
\setlength{\tabcolsep}{4.pt}
\footnotesize
\begin{tabular}{l|l|cccccc}
\toprule
\textbf{} & \textbf{Method} & \textbf{KonIQ} & \textbf{SPAQ} & \textbf{KADID} \\
\midrule
1
& Q-Align & 0.943 / 0.940 & 0.933 / 0.931 & 0.692 / 0.708 \\
& DeQA-Score & 0.953 / 0.943 & 0.936 / 0.933 & \textbf{0.724 / 0.719}  \\
\rowcolor{gray!30} 
& \textbf{Ours (5)} & 0.954 / 0.941 & 0.936 / 0.932 & 0.672 / 0.661 \\
\rowcolor{gray!50} 
& \textbf{Ours (1)} & 0.950 / 0.936 & 0.934 / 0.931 & 0.646 / 0.628 \\
\rowcolor{gray!70} 
& \textbf{Ours (F-loss)} & \textbf{0.961 / 0.950} & \textbf{0.937 / 0.934} & 0.702 / 0.689 \\
\midrule
2
& Q-Align & 0.945 / 0.938 & 0.933 / 0.931 & 0.935 / 0.934 \\
& DeQA-Score & 0.957 / 0.944 & \textbf{0.938} / 0.934 & 0.955 / 0.953 \\
\rowcolor{gray!30} 
& \textbf{Ours (5)} & 0.953 / 0.939 & 0.936 / 0.932 & 0.961 / 0.958 \\
\rowcolor{gray!50} 
& \textbf{Ours (1)} & 0.952 / 0.938 & 0.934 / 0.930 & 0.959 / 0.956 \\
\rowcolor{gray!70} 
& \textbf{Ours (F-loss)} & \textbf{0.961 / 0.951} & \textbf{0.938 / 0.935} & \textbf{0.963 / 0.961} \\
\bottomrule
\end{tabular}
\caption{Comparison of multi-dataset training results across different IQA datasets. “1” denotes training on KonIQ and SPAQ; “2” denotes training on KonIQ, SPAQ and KADID. “F-loss” denotes Fidelity loss. (PLCC / SRCC)}
\label{tab3}
\vspace{-0.4cm}
\end{table}

\section{Experiments}
In this section, to demonstrate the effectiveness and generalization ability of Q-Scorer, we conduct experiments on both single-dataset settings (Sec.~\ref{single-dataset}) and multi-dataset settings (Sec.~\ref{multi-dataset}). In addition, we carry out ablation studies to verify the effectiveness of the two key components in Q-Scorer (Sec.~\ref{ablation}). To further validate the efficiency of the score token embedding, we perform comprehensive comparisons with the VideoScore method from multiple perspectives (Sec.~\ref{efficiency}). Finally, we explore the integration of Q-Scorer with other methods and observe its compatibility and extensibility (Sec.~\ref{combination}).

\subsection{Experimental Settings}
We fine-tune our model using the LoRA strategy~\cite{hu2022lora}, starting from the pre-trained mPLUG-Owl2~\cite{ye2024mplug} weights. The visual encoder is initialized with CLIP-pretrained ViT-L~\cite{radford2021learning}, and the language backbone is LLaMA-2-7B~\cite{touvron2023llama}. Training is performed for 3 epochs with a batch size of 16 using the AdamW optimizer~\cite{loshchilov2017decoupled}, an initial learning rate of 2e-5, and cosine decay scheduling. All experiments run on 2 NVIDIA RTX A100 GPUs, and training on KonIQ completes in about 30 minutes.

\begin{table*}[t]
\centering
\setlength{\tabcolsep}{4.pt}
\footnotesize
\begin{tabular}{l|l|c|ccccc}
\toprule
\textbf{Type} & \textbf{Ablation} & \textbf{KonIQ} & \textbf{SPAQ} & \textbf{KADID}  & \textbf{LIVE-Wild} & \textbf{AGIQA-3K} & \textbf{CSIQ} \\
\midrule
Regression module
& without & 0.944 / 0.942 & 0.887 / 0.893 & 0.629 / 0.622 & 0.871 / 0.855 & 0.816 / 0.735 & 0.755 / 0.717 \\
& [512, 256] & 0.957 / 0.948 & 0.895 / 0.893 & 0.643 / 0.629 & 0.883 / 0.869 & 0.796 / 0.706 & 0.782 / 0.731 \\
& [2048, 1024, 512, 256] & 0.938 / 0.927 & 0.869 / 0.866 & 0.586 / 0.576 & 0.843 / 0.844 & 0.810 / 0.689 & 0.743 / 0.697 \\
\midrule
Score token
& level token (mix) & 0.948 / 0.940 & 0.896 / 0.896 & 0.653 / 0.635 & 0.873 / 0.853 & 0.772 / 0.685 & 0.759 / 0.684 \\
& level token (bare) & 0.946 / 0.935 & 0.887 / 0.883 & 0.607 / 0.592 & 0.862 / 0.854 & 0.817 / 0.707 & 0.774 / 0.727 \\
\midrule
\rowcolor{gray!40} 
& \textbf{Our (base)} & \textbf{0.959 / 0.948} & \textbf{0.898 / 0.898} & \textbf{0.676 / 0.671} & \textbf{0.889 / 0.870} & \textbf{0.821 / 0.734} & \textbf{0.796 / 0.732} \\
\bottomrule
\end{tabular}
\caption{Ablation study results. “without” denotes using weighted sum instead of an MLP. “[*]” indicates different MLP settings. “mix” uses both level and score tokens; “bare” uses only level token embeddings for regression. (PLCC / SRCC)}
\label{tab4}
\vspace{-0.4cm}
\end{table*}

\subsection{Datasets and Baselines}
We train our model on three IQA datasets: KonIQ~\cite{hosu2020koniq}, SPAQ~\cite{fang2020perceptual}, and KADID~\cite{lin2019kadid}, following the setup in Q-Align. To assess generalization, we evaluate on four unseen datasets: LIVE-Wild~\cite{ghadiyaram2015live}, AGIQA-3K~\cite{li2023agiqa}, and CSIQ~\cite{larson2010most}. The MOSs of these datasets are normalized to $[1,5]$.

We primarily compare against Q-Align~\cite{wu2023qa}, its LoRA-adapted variant, and its improved method DeQA-Score~\cite{you2025teaching}. Besides, we also report results from representative traditional and deep learning-based methods, including handcrafted metrics (NIQE~\cite{mittal2012making}, BRISQUE~\cite{mittal2012no}) and learning-based models (NIMA~\cite{talebi2018nima}, HyperIQA~\cite{su2020blindly}, DBCNN~\cite{zhang2018blind}, MUSIQ~\cite{ke2021musiq}, CLIP-IQA+\cite{wang2023exploring}, ManIQA\cite{yang2022maniqa}, and Compare2Score~\cite{zhu2024adaptive}).

\subsection{Metrics}
We use the Pearson Linear Correlation Coefficient (PLCC) and Spearman Rank-order Correlation Coefficient (SRCC) to evaluate score regression performance. PLCC assesses the linear correlation between predicted scores and ground-truth MOSs, while SRCC evaluates the consistency of their ranking order.

\subsection{Single-Dataset Training Results}
\label{single-dataset}
We first train our model solely on the KonIQ dataset and evaluate its generalization on all other datasets, as summarized in Tab.\ref{tab2}. To fairly assess LoRA-based fine-tuning, we include a LoRA-adapted Q-Align as a baseline. Our method consistently outperforms baselines in score regression across all datasets except KADID, demonstrating strong generalization. On the KonIQ, it achieves state-of-the-art results, improving PLCC by 2\% over fully fine-tuned Q-Align, and by 2.9\% over its LoRA-adapted variant. The performance drop on KADID likely stems from LoRA’s side effects, as Q-Align with LoRA shows a similar decline. Interestingly, using a single score token outperforms the enhanced token set here, possibly due to the smaller data scale and simpler score distribution of a single dataset, where one token is sufficient for effective mapping.

\subsection{Multi-Dataset Training Results}
\label{multi-dataset}
We present multi-dataset co-training results in Tab.\ref{tab3}, where our method consistently outperforms Q-Align across various dataset combinations. However, when trained on all three datasets simultaneously, our model lags behind the DeQA-Score baseline, which leverages fidelity loss to improve cross-dataset generalization. This is expected, as different IQA datasets often exhibit domain shifts, where the same MOS may represent different perceptual qualities~\cite{zhu2024adaptive}. Using a unified set of score tokens and a single MLP head to represent all datasets may cause interference, especially reflected in the KonIQ performance drop. By simply adding fidelity loss, our method achieves state-of-the-art results (see the last row of Tab.\ref{tab3}; details in the Appendix). Additionally, relying on a single score token significantly degrades performance in multi-dataset settings, underscoring the necessity of adopting a set of interval-specific score tokens to better model challenging datasets.

\subsection{Ablation Studies}
\label{ablation}
As shown in Tab.\ref{tab4}, we conduct ablation studies to evaluate the impact of the regression module and score token design. Our proposed configuration consistently achieves the best performance. Notably, replacing the MLP with a probability-weighted sum over score tokens results in a significant performance drop, demonstrating that regression-based modeling is not only theoretically optimal but also empirically effective. Moreover, mixing level tokens with score tokens or directly using level token embeddings for regression degrades performance, further highlighting the semantic confusion introduced by level tokens.

\subsection{Score Token Embedding Efficiency}
\label{efficiency}
RealQA~\cite{li2025tokenenoughrealisticimage} notes that while existing regression-based method (VideoScore) improves with more training, they still lag behind next-token prediction (NTP) approaches. This is mainly due to the insufficient embedding expressiveness of the last token to model score distributions, leading to slow convergence and degraded performance. As shown in Tab.\ref{tab5}, our model alleviates these issues through a dedicated score token design, achieving accurate score prediction within 1 epoch and full convergence in 3–4 epochs.

\subsection{Combination with Other Methods}
\label{combination}
Our method can be further improved by integrating existing techniques. To mitigate potential overlap among score tokens, we incorporate the KL divergence loss from DeQA-Score. We also explore enhancements with external modules such as norm-in-norm loss~\cite{peng2023dfgc}, hyper network~\cite{su2020blindly} and ranking loss~\cite{liu2017rankiqa}. These combinations yield additional gains on several datasets, highlighting the compatibility and extensibility of our approach (details in the Appendix).

\begin{table}[t]
\footnotesize
\centering
\label{tab:training-efficiency}
\begin{tabular}{l|c|c}
\toprule
\textbf{Method} & \textbf{Epochs} & \textbf{KonIQ} \\
\midrule
VideoScore & 2 & 0.890 / 0.882 \\
VideoScore & 6 & 0.923 / 0.908 \\
\rowcolor{gray!30} 
Ours & 1 & 0.929 / 0.919 \\
\rowcolor{gray!30} 
Ours & 2 & 0.957 / 0.945 \\
\rowcolor{gray!70} 
Ours (base) & 3 & \textbf{0.959} / 0.948 \\
\rowcolor{gray!70} 
Ours & 4 & \textbf{0.959 / 0.949} \\
\rowcolor{gray!50} 
Ours & 5 & 0.957 / 0.947 \\
\rowcolor{gray!50} 
Ours & 6 & 0.957 / 0.948 \\
\bottomrule
\end{tabular}
\caption{Performance comparison under different training epochs and methods. (PLCC / SRCC)}
\label{tab5}
\vspace{-0.4cm}
\end{table}

\section{Conclusions}
Our work identifies conversion errors and semantic confusion in existing MLLM-based IQA methods, and introduce Q-Scorer, a simple yet effective framework that leverages IQA-specific score tokens and a lightweight regression module to predict continuous quality scores. Q-Scorer achieves state-of-the-art performance across multiple IQA benchmarks, offering a promising direction for enhancing the quality assessment capabilities of MLLMs.

\noindent \textbf{Limitations and Future Work.} Most IQA datasets primarily reflect annotators’ personal preferences, so models trained on their diverse distributions of their MOSs may fail to capture how people in general perceive image quality. In future work, we plan to incorporate self-learning mechanisms into MLLMs to improve their generalization beyond dataset-specific biases.

\section{Acknowledgments}
This work is supported by the National Natural Science Foundation of China (NSFC) under Grant No. 62272460, Beijing Natural Science Foundation under Grant No. 4232037

\bibliography{aaai2026}

\newpage

\appendix

{\centering\Large\bfseries Appendix \par}

\section{Overview}
This appendix is structured as follows. Appendix~\ref{proof} provides additional proof details. Appendix~\ref{result} provides experimental details and extra qualitative results. Finally, Appendix~\ref{extension} introduces some possible extensions.

\section{Proofs}
\label{proof}

\subsection{Q-Align}
For Q-Align, the one-sample error $\epsilon(x)$ varies across different discretization intervals. For example, when the sign of the error is taken into account, in the interval $[1, 1.8]$, the discretized label is chosen as $1$, and we have $\epsilon(x) \in [0, 0.8]$. Similarly, in the interval $[1.8, 2.6]$ where the discretized value is $2$, the error becomes $\epsilon(x)\in [-0.2, 0.6]$.

Assuming $\epsilon(x)$ follows a uniform distribution over each discretization interval, the expected squared error can be calculated as:
\begin{equation}
\begin{aligned}
\mathrm{E}[\epsilon(x)^2] 
&= \sum_{i=1}^{5} \int_a^b \frac{1}{b - a} \epsilon(x)^2 \, d\epsilon(x) \\
&= \frac{1}{5} \Biggl(
    \int_0^{0.8} \frac{1}{0.8} \epsilon^2 \, d\epsilon 
  + \int_{-0.2}^{0.6} \frac{1}{0.8} \epsilon^2 \, d\epsilon \\
&\quad
  + \int_{-0.4}^{0.4} \frac{1}{0.8} \epsilon^2 \, d\epsilon 
  + \int_{-0.6}^{0.2} \frac{1}{0.8} \epsilon^2 \, d\epsilon \\
&\quad
  + \int_{-0.8}^{0} \frac{1}{0.8} \epsilon^2 \, d\epsilon
\Biggr) \\
&= \frac{18}{125} > 0.
\end{aligned}
\end{equation}
where $[a, b]$ denotes the error interval.

\subsection{DeQA-Score}

As discussed in the main text, DeQA-Score exhibits a label approximation error $\epsilon_2(x)$ when the enhancement module is not applied:
\begin{equation}
\epsilon_2(x) = \left| \hat{y}(x) - S \right| 
= \left| \sum_{i=1}^{5} c_i p_i^{raw} - \int_{-\infty}^{\infty} sf(s)\, ds \right|.
\end{equation}

Here, the discrete form $\sum_{i=1}^{5} c_i p_i^{raw}$ can be viewed as a midpoint-based approximation of the truncated expectation:
\begin{equation}
\sum_{i=1}^{5} c_i p_i^{raw} =\sum_{i=1}^{5} c_i \cdot \int_{c_i - \frac{1}{2}}^{c_i + \frac{1}{2}} f(s)\, ds \approx \int_{0.5}^{5.5} sf(s)\, ds,
\end{equation}
Letting $h(s) = sf(s)$, the per-interval approximation error $E_i$ is:
\begin{equation}
E_i = -\frac{1}{24} \cdot h''(\xi_i),
\quad \xi_i \in [c_i - 0.5, c_i + 0.5],
\end{equation}

Assuming the second derivative of $h(x)$ is bounded over $[0.5, 5.5]$,
\begin{equation}
|h''(x)| \leq M, \quad \forall x \in [0.5, 5.5]
\end{equation}
the total label approximation error is therefore bounded:
\begin{equation}
\left| \sum_{i=1}^{5} c_i p_i^{raw} - \int_{0.5}^{5.5} x f(x)\, dx \right| \leq \frac{5}{24} M.
\end{equation}

Furthermore, due to the tail truncation of the Gaussian distribution:
\begin{equation}
\int_{0.5}^{5.5} x f(x)\, dx < \int_{-\infty}^{\infty} x f(x)\, dx = s,
\end{equation}
we conclude that $\hat{y}(x)$ almost never equals the ground-truth quality score $s$, thus:
\begin{equation}
\epsilon_2(x) > 0.
\end{equation}

\subsection{Regression-Based Modeling}
We provide a constructive proof for the Universal Approximation Theorem (UAT) used in the main text. Let $\mathcal{F}(x)$ denote the score prediction function realized by a neural network, and $g(x)$ denote the target function that maps the input $x \in [0,1]^n$ to the corresponding MOS $S$. The goal is to show that for any $\varepsilon > 0$, there exists a neural network $\mathcal{F}$ such that:
\begin{equation}
\sup_{x \in [0,1]^n} \left| \mathcal{F}(x) - g(x) \right| < \varepsilon.
\end{equation}

To quantify the approximation, consider the mean squared error over the domain:
\begin{equation}
E = \int_{[0,1]^n} \left( \mathcal{F}(x) - g(x) \right)^2 dx.
\end{equation}

Assume the activation function $\sigma(x)$ is continuous and bounded. According to the Weierstrass approximation theorem, for any $\delta > 0$, there exists a polynomial $p(x)$ such that
\begin{equation}
|\sigma(x) - p(x)| < \delta, \quad \forall x \in \mathrm{R}.
\end{equation}

Using this polynomial to approximate the activation function, the neural network can be expressed in the following form:
\begin{equation}
\mathcal{F}(x) = \sum_{i=1}^{n} \alpha_i \cdot p(w_i^\top x + b_i),
\end{equation}

Substituting this into the error expression gives:
\begin{equation}
E = \int_{[0,1]^n} \left( \sum_{i=1}^{n} \alpha_i \cdot p(w_i^\top x + b_i) - g(x) \right)^2 dx.
\end{equation}

Since polynomials are dense in the space of continuous functions on compact domains with respect to the uniform norm, there exists a finite linear combination of terms \( p(w_i^\top x + b_i) \) that can approximate \( g(x) \) arbitrarily well (Weierstrass approximation theorem). Consequently, the function \( \mathcal{F}(x) \) uniformly approximates \( g(x) \) on the domain \([0,1]^n\), thus completing the proof.

\section{Results}
\label{result}

\begin{table*}[htbp]
\centering
\footnotesize
\begin{tabular}{l|c|ccccc}
\toprule
\textbf{Methods} & \textbf{KonIQ} & \textbf{SPAQ$^{cr}$} & \textbf{KADID$^{cr}$}  & \textbf{LIVE-Wild$^{cr}$} & \textbf{AGIQA-3K$^{cr}$} & \textbf{CSIQ$^{cr}$} \\
\midrule

Q-Align & 0.941 / 0.940 & 0.886 / 0.887 & 0.674 / 0.684 & 0.853 / 0.860 & 0.772 / 0.735 & 0.785 / 0.737 \\
DeQA-Score & 0.953 / 0.941 & 0.895 / 0.896 & \textbf{0.694 / 0.687} & 0.892 / 0.879 & 0.809 / 0.729 & 0.787 / 0.744 \\
Q-Align (LoRA) & 0.932 / 0.938 & 0.874 / 0.886 & 0.624 / 0.632 & 0.858 / 0.859 & 0.806 / 0.735 & 0.772 / 0.730 \\
\textbf{Ours (5)} & 0.959 / 0.948 & 0.898 / 0.898 & 0.676 / 0.671 & 0.889 / 0.870 & 0.821 / 0.736 & 0.796 / 0.746 \\
\textbf{Ours (1)} & 0.960 / 0.950 & \textbf{0.900 / 0.899} & 0.660 / 0.645 & \textbf{0.903 / 0.888} & 0.811 / 0.722 & 0.795 / 0.733 \\
\textbf{Ours (KL)} & \textbf{0.963 / 0.953} & 0.898 / 0.898 & 0.643 / 0.630 & 0.881 / 0.858 & 0.786 / 0.713 & 0.778 / 0.734 \\
\textbf{Ours (N-loss)} & 0.959 / 0.948 & 0.896 / 0.896 & 0.679 / 0.670 & 0.889 / 0.871 & 0.819 / 0.737 & 0.804 / 0.746 \\
\textbf{Ours (Hyper16)} & 0.950 / 0.940 & 0.894 / 0.896 & 0.643 / 0.624 & 0.860 / 0.841 & 0.821 / 0.760 & 0.778 / 0.737 \\
\textbf{Ours (Hyper64)} & 0.946 / 0.930 & 0.893 / 0.893 & 0.652 / 0.629 & 0.853 / 0.822 & \textbf{0.823 / 0.762} & \textbf{0.827 / 0.783} \\
\textbf{Ours (R-loss)} & 0.961 / 0.951 & 0.895 / 0.895 & 0.657 / 0.642 & 0.895 / 0.882 & 0.813 / 0.711 & 0.792 / 0.731 \\
\bottomrule
\end{tabular}
\caption{Performance comparison across multiple IQA datasets (trained on KonIQ). “cr” denotes cross-dataset evaluation. “Ours(1)” refers to predicting the same score token regardless of input. “Ours(5)” denotes our base method. “Ours(N-loss)” introduces an additional norm-in-norm loss. “Ours(Hyper16)” and “Ours(Hyper64)” denote the use of hyper networks with hidden dimensions of 16 and 64, respectively. “Ours(R-loss)” introduces an additional ranking loss. (PLCC / SRCC)}
\label{tab}
\end{table*}

\subsection{Multi-Dataset Training}
To address the domain shifts among different IQA datasets, we apply fidelity loss to enhance the model's generalization capability. For each image pair $(x, y)$, fidelity loss leverages the relative ranking information inferred from the corresponding MOS values and their variances. Denoting the annotated mean (MOS) and variance of two images as $\mu_x$, $(\sigma_x)^2$ and $\mu_y$, $(\sigma_y)^2$, we estimate the probability that image $x$ is perceptually better than image $y$ as:
\begin{equation}
p(x > y) = \Phi \left( \frac{\mu_x - \mu_y}{\sqrt{(\sigma_x)^2 + (\sigma_y)^2}} \right),
\end{equation}
where $\Phi(\cdot)$ is the cumulative distribution function of the standard normal distribution. Meanwhile, the model predicts $\mu_x^{pred}$, $(\sigma_x^{pred})^2$ and $\mu_y^{pred}$, $(\sigma_y^{pred})^2$. The predicted probability is:
\begin{equation}
p^{pred}(x > y) = \Phi \left( \frac{\mu_x^{pred} - \mu_y^{pred}}{\sqrt{(\sigma_x^{pred})^2 + (\sigma_y^{pred})^2}} \right),
\end{equation}
Then, the fidelity loss is calculated to measure the similarity between predicted and ground-truth preferences:

\begin{equation}
\begin{aligned}
    \mathcal{L}_{fd} = 1 
    &- \sqrt{p(x>y) \cdot p^{pred}(x>y)} \\
    &- \sqrt{(1 - p(x>y)) \cdot (1 - p^{pred}(x>y))}.
\end{aligned}
\end{equation}

Due to the lack of variance prediction in Q-Scorer, we briefly explore two approximation methods: (1) a lightweight MLP head (2-dim output) that directly regresses the mean and variance from score tokens, and (2) the DeQA-Score strategy, which infers variance from the score token distribution. Both methods achieve state-of-the-art performance, with the second method performing slightly better. This is because score tokens are enhanced by intervals and only implicitly encode variance in their distribution, so directly regressing variance from the output score token tends to introduce larger errors. However, the token-distribution-based method also has limited accuracy due to its coarse weighted scheme. As in the original BIQA paper, a simple MLP remains an effective choice for predicting variance. Future work could explore score token designs that better exploit or explicitly encode variance information.

\subsection{Combination with Other Methods}
As shown in Tab.~\ref{tab}, our method achieves additional gains on several datasets by incorporating one of the following: the KL divergence loss from DeQA-Score, norm-in-norm loss, hyper network, or ranking loss.

\subsubsection{KL Divergence Loss.}
Although we divide the score tokens according to predefined intervals, and each score token is theoretically expected to predict scores within its corresponding interval, in practice, the LLM often introduces bias when predicting scores, especially for values near the interval boundaries. This leads to shifts in the effective score range each token represents, causing unintended overlaps between score tokens. To mitigate this potential overlap, we incorporate the KL divergence loss from DeQA-Score.

Specifically, we adopt the soft label strategy from DeQA-Score, where the enhanced soft label is $P$ and the predicted discrete distribution is $P_p$ (computed by applying a softmax to the output logits). The KL divergence loss is then formulated as:
\begin{equation}
\mathcal{L}_{KL} = \sum_i p_i \log \left( \frac{p_i}{p_i^{pred}} \right).
\label{eq:kl}
\end{equation}

This loss effectively enhances model performance on the in-distribution KonIQ dataset by aligning the predicted distribution with soft targets and reducing score-token overlap, although it may slightly compromise generalization on other out-of-distribution datasets.

\subsubsection{Norm-in-norm Loss.}

To improve generalization, we incorporate an auxiliary norm-in-norm loss with weight of 0.5 in the total loss computation. Specifically, the norm-in-norm loss normalizes both ground-truth and predicted scores to encourage linear relationships and accelerate convergence. Given ground-truth scores $S$ and predicted scores $\hat{y}(x)$, the loss is defined as:

\begin{equation}
\mathcal{L}_{NIN}(S, \hat{y}(x)) = \sum_{i=1}^N \left| Q_i - \hat{Q}_i \right|,
\end{equation}

where $Q_i$ is the normalized version of $S_i$:

\begin{equation}
Q_i = \frac{S_i - \frac{1}{N}\sum_{j=1}^N S_j}{\left( \sum_{j=1}^N \left| S_j - \frac{1}{N}\sum_{k=1}^N S_k \right|^q \right)^{1/q}}.
\end{equation}
and $\hat{Q}_i$ is computed similarly from $\hat{y}(x)$. The hyperparameter $q$ is set to $2$. 

This loss demonstrates superior performance by enhancing robustness on out-of-distribution datasets while preserving performance on in-distribution dataset.

\subsubsection{Hyper Network.}
To cover wide content variation and enhance the generalization capability of the model, we follow the design in HyperIQA by using a hyper network to adaptively learn the rule for perceiving quality based on the recognized image content. Specifically, we introduce a hyper network $H(\cdot)$ to simplify the prediction problem:

\begin{equation}
    \theta_x = H(G(x), \gamma),
\end{equation}
where $H$ denotes the hyper network mapping function (an MLP that maps from semantic feature dimensions to the number of parameters), and $\gamma$ represents its parameters. $G(x)$ denotes the semantic feature extracted from the input image $x$, which we simplify as the score token embedding in our framework. Thus, the hyper network learns a mapping from image content to quality prediction rules. The final prediction by the target network can be formulated as:

\begin{equation}
    F(z, H(G(x), \gamma)) = \hat{y}(x).
\end{equation}
where $z$ is the visual representation (score token embedding) of the image $x$, and $\hat{y}(x)$ is the predicted quality score.

Although this design reduces performance on the in-distribution KonIQ dataset, it significantly improves generalization in out-of-distribution datasets like AGIQA-3K and CSIQ. However, due to limited computational resources, we restrict the hyper network to 64 hidden units and adjust the target network $F$ from the original structure $[2048, 1024]$ to a more lightweight $[512, 256]$ configuration. Meanwhile, different strategies for selecting the input embedding (e.g., combining compressed visual token embeddings) also influence the performance. These findings indicate the potential for future exploration.

\subsubsection{Ranking Loss.}

To better model relative perceptual quality, we simplify the Siamese network and incorporate a ranking loss, inspired by RankIQA. Specifically, for each image $x_i$, we extract a compact semantic representation $z_i$ (i.e., score token embedding). Given a pair $(x_i, x_j)$ sampled from the mini-batch such that $x_i > x_j$ (i.e., $x_i$ has higher MOS), both embeddings $z_i$ and $z_j$ are passed through a shared MLP scoring network $F$, yielding predicted scores $\hat{y}_i = F(z_i)$ and $\hat{y}_j = F(z_j)$. A pairwise ranking loss encourages correct quality ordering:
\begin{equation}
    \mathcal{L}_{\text{rank}} = \max(0, \hat{y}_j - \hat{y}_i + \epsilon),
\end{equation}
where $\epsilon$ is a margin hyperparameter. However, unlike the original RankIQA framework that trains ranking loss and score loss in two stages, we jointly optimize both ranking and regression objectives during training.

Although the introduction of the ranking loss does not lead to significant improvements on both in-distribution and out-of-distribution datasets compared to other combinations, we argue that it remains a worthwhile direction for exploration due to its conceptual simplicity and the limitations imposed by small batch sizes under resource constraints.

\subsection{Qualitative Results}
\begin{figure*}[t]
\centering
\begin{subfigure}{0.7\textwidth}
    \centering
    \includegraphics[width=\linewidth]{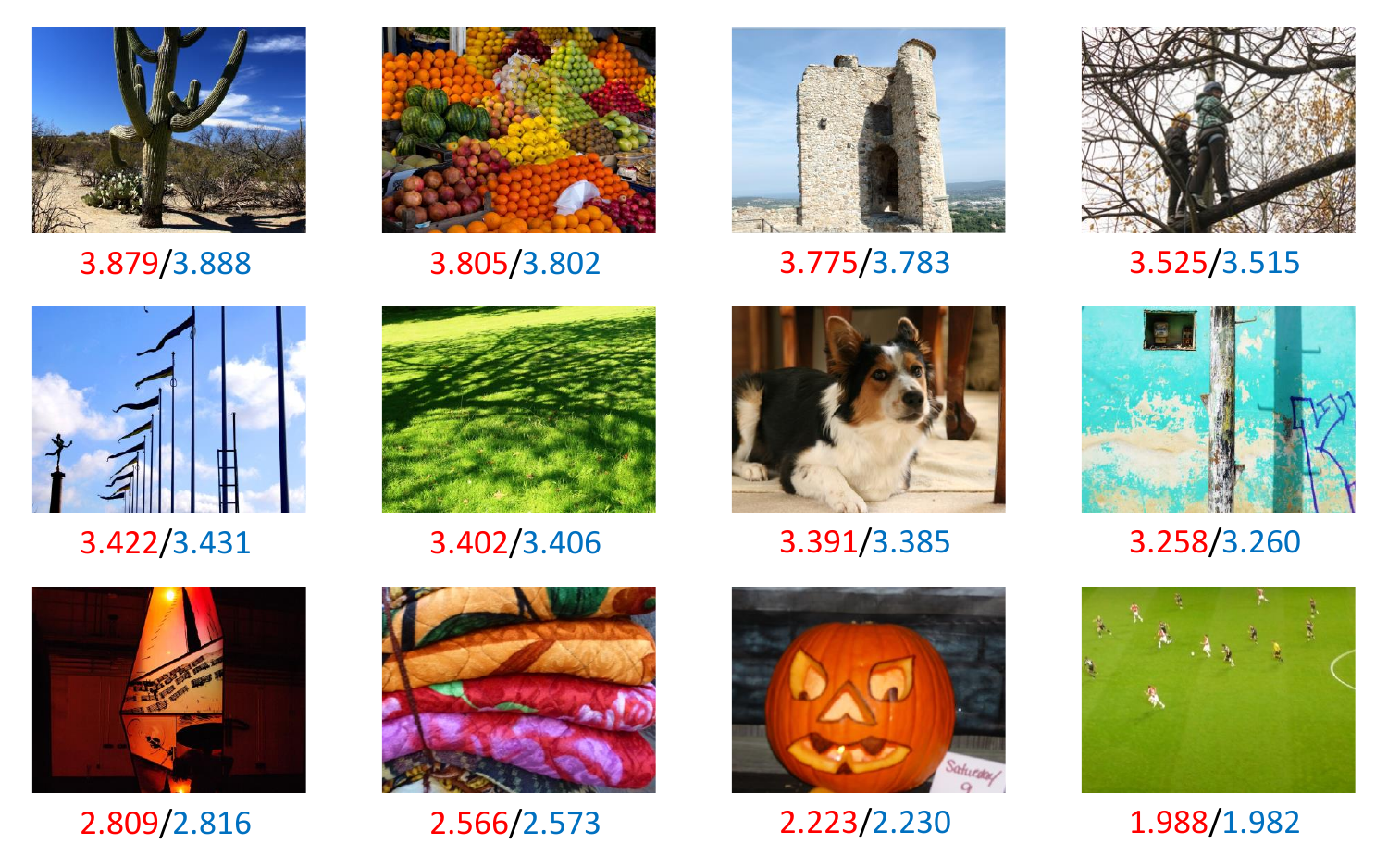}
    \caption{Qualitative results on in-the-wild IQA datasets.}
    \label{q1}
\end{subfigure}

\vspace{-0.3em}

\begin{subfigure}{0.7\textwidth}
    \centering
    \includegraphics[width=\linewidth]{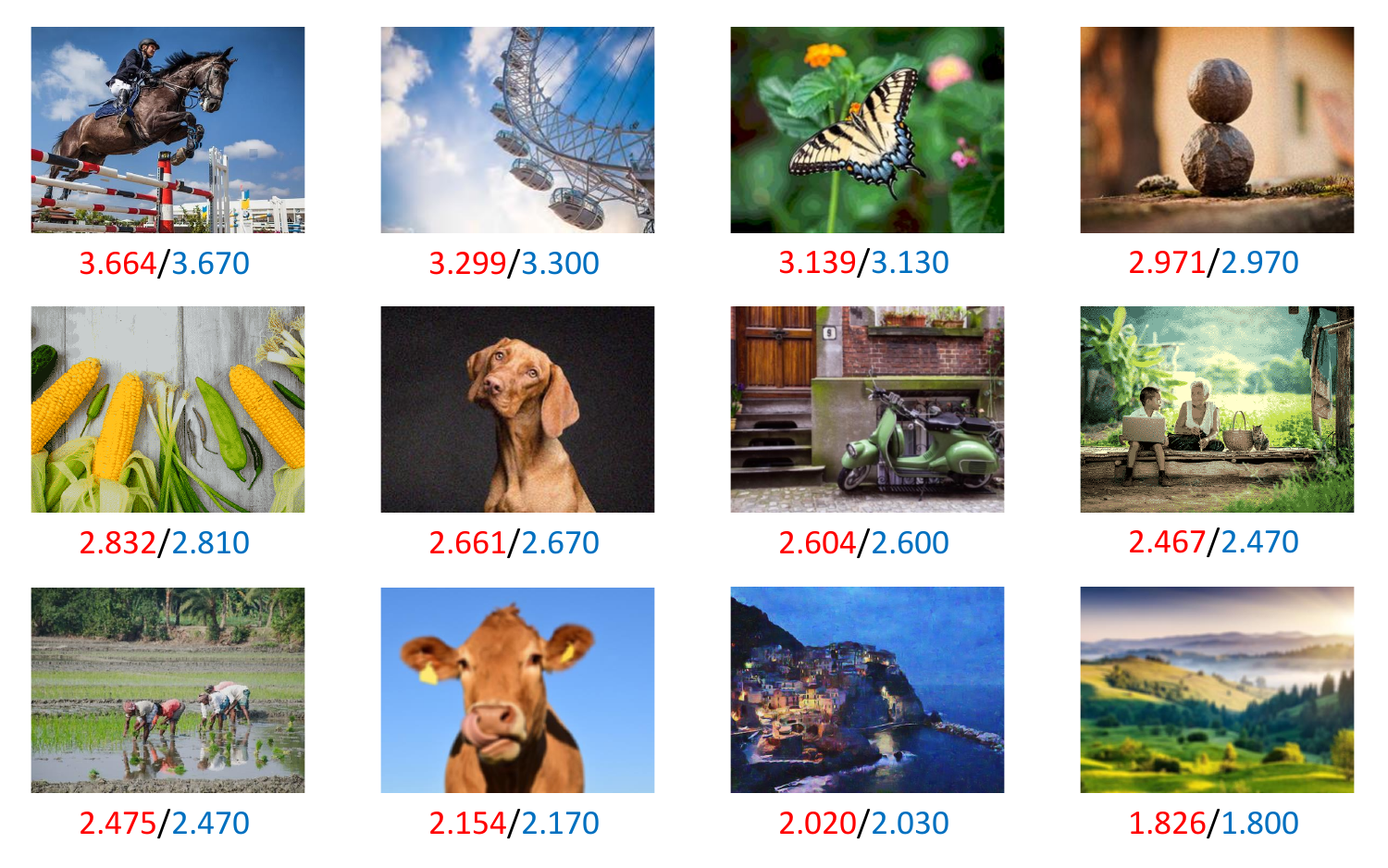}
    \caption{Qualitative results on synthetically distorted IQA datasets.}
    \label{q2}
\end{subfigure}

\vspace{0.3em}

\begin{subfigure}{0.7\textwidth}
    \centering
    \includegraphics[width=\linewidth]{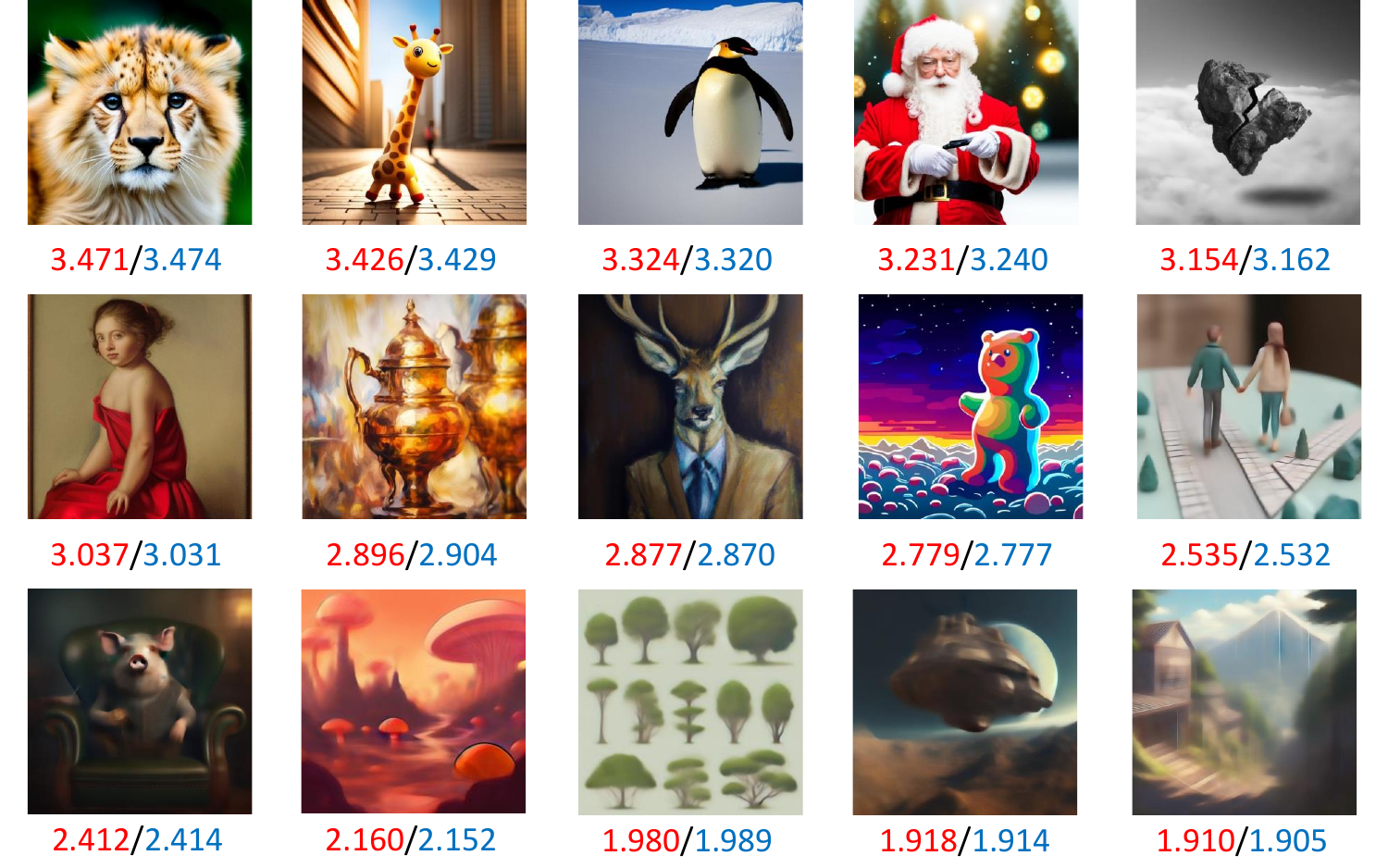}
    \caption{Qualitative results on AI-generated IQA datasets.}
    \label{q3}
\end{subfigure}

\vspace{-0.8em}

\caption{Qualitative results across three types of IQA datasets. Predicted scores are shown in red, and ground-truth MOSs are shown in blue.}
\label{fig:qualitative_all}
\end{figure*}

We present several qualitative results in Fig.\ref{q1}, Fig.\ref{q2}, and Fig.~\ref{q3}, covering diverse image types including in-the-wild images (KonIQ, SPAQ, and LIVE-Wild datasets), synthetically distorted images (KADID and CSIQ datasets), and AI-generated images (AGIQA-3K dataset), as well as a wide range of image qualities and varied content. These results demonstrate that our method produces quality assessments that closely align with human evaluations.

\section{Extensions}
\label{extension}

\subsection{Improving Score Token Alignment}
We observe that when using only cross-entropy loss $\mathcal{L}_{ce}$ to guide the LLM in generating appropriate tokens, the cumulative probability of level tokens in token-based methods often approaches 1, indicating good alignment. However, although the model generally predicts the correct score token, the cumulative probability assigned to score tokens frequently falls below 1. This suggests suboptimal alignment of score tokens, likely due to the fact that these newly introduced tokens were not part of the original vocabulary and hence lack grounding across general pretraining tasks. This phenomenon reflects a limitation in score token alignment and highlights the potential for further refinement. A potential remedy is to introduce an auxiliary loss that encourages the cumulative probability over the score tokens to approach 1. Exploring such score-calibrated objectives offers a promising direction for further enhancing the effectiveness of score tokens.

\subsection{Integrating Traditional IQA Methods}

Although experiments demonstrate that combining Q-Scorer with other methods yields performance gains on several datasets, there remains considerable potential to be unlocked. For instance, employing larger hidden dimensions in hyper network or using larger batch sizes in ranking loss may further boost performance. However, due to current limitations in computational resources and time, we have only conducted preliminary trials. More importantly, Q-Scorer effectively integrates a regression module into MLLMs via IQA-specific tokens, paving the way for adapting more classical IQA methods, most of which are regression-based, into the MLLM framework in future research.

\end{document}